%% file: main.tex
\documentclass[letterpaper, 10 pt, conference]{ieeeconf}
\IEEEoverridecommandlockouts

\usepackage{cite}
\usepackage{amsmath,amssymb,amsfonts}
\usepackage{algorithm,algorithmicx,algpseudocode}
\usepackage{graphicx}
\usepackage{textcomp}
\usepackage{xcolor}

\usepackage{pgfplots}
\usepackage{diagbox}
\usepackage{multirow}
\usepackage{hyperref}
\usepackage{longtable}

\def\BibTeX,{{\rm B\kern-.05em{\sc i\kern-.025em b}\kern-.08em
    T\kern-.1667em\lower.7ex\hbox{E}\kern-.125emX}}
\begin{document}

\bibliographystyle{IEEEtran}

\title{\LARGE \bf
Harmonizing Human Insights and AI Precision: \\ Hand in Hand for Advancing Knowledge Graph Task
}


\author{
Shurong Wang$^{1}$, Yufei Zhang$^{2}$, Xuliang Huang$^{1}$, and Hongwei Wang$^{1*}$%
\thanks{$^{1}$ ZJU-UIUC Institute, Zhejiang University, Haining, China}%
\thanks{$^{2}$ College of Biomedical Engineering and Instrument Science, Zhejiang University, Hangzhou, China}%
\thanks{$^{*}$ Corresponding author, email \texttt{hongweiwang@zju.edu.cn}}
}

\maketitle

\begin{abstract}
Knowledge graph embedding (KGE) has caught significant interest for its effectiveness in knowledge graph completion (KGC), specifically link prediction (LP), with recent KGE models cracking the LP benchmarks. Despite the rapidly growing literature, insufficient attention has been paid to the cooperation between humans and AI on KG.
However, humans' capability to analyze graphs conceptually may further improve the efficacy of KGE models with semantic information. To this effect, we carefully designed a human-AI team (HAIT) system dubbed KG-HAIT, which harnesses the human insights on KG by leveraging fully human-designed ad-hoc dynamic programming (DP) on KG to produce human insightful feature (HIF) vectors that capture the subgraph structural feature and semantic similarities. By integrating HIF vectors into the training of KGE models, notable improvements are observed across various benchmarks and metrics, accompanied by accelerated model convergence. Our results underscore the effectiveness of human-designed DP in the task of LP, emphasizing the pivotal role of collaboration between humans and AI on KG. We open avenues for further exploration and innovation through KG-HAIT, paving the way towards more effective and insightful KG analysis techniques.

\end{abstract}



\section{Introduction}



Knowledge Graphs (KGs) take advantage of the expressiveness of heterogeneous graphs to support applications in diverse fields such as relation finding, question answering, and recommendation systems. Despite their immense scale, KGs suffer from inherent incompleteness \cite{KG-Incomplete} since manually collecting knowledge can never be sufficient. Hence, strategies for mining additional facts to augment KGs known as knowledge graph completion (KGC) \cite{KGC} has sparked considerable interest.
Link prediction (LP), a core solution to KGC, is dedicated to extracting new and confident knowledge from existing knowledge in KGs. It has gained increasing attention and been largely fueled by machine learning (ML) techniques \cite{TransE}, \cite{NBFNet}.

\begin{figure}[!t]
    \centering
    \includegraphics[width = 0.5\textwidth]{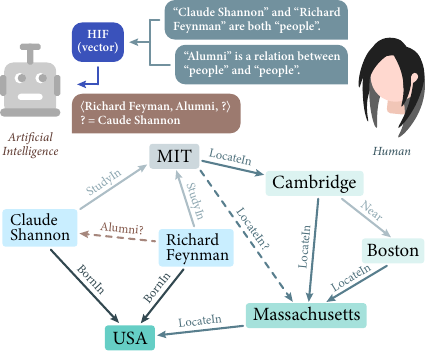}
    \caption{Subgraph of a knowledge graph with actual relations between entities (solid lines) and inferred relations (dashed lines). Entities with the same type are presented with the same color.}
    \label{fig:KG}
\end{figure}



To automatically predict missing links between entities based on known ones accomplishing the objective of LP, knowledge graph embedding (KGE) is introduced. Inspired by word embedding \cite{Word-Embedding}, KGE represents entities and relations in KGs as low dimensional vectors or tensors, and this transition from discrete to continuous enables the involvement of ML methods. The employment of ML has proven promising as numerous recent KGE models have been leading across LP benchmarks \cite{KGE_Summary}. Although AI (KGE models) with extreme computational power and high precision outperform humans in LP, they can't conceptually analyze KGs and discover crucial abstractions as humans. These disabilities make it difficult for KGE models to score higher in benchmarks, as they have taken full advantage of fitting the training data via the training algorithm.

However, most existing LP methods heavily rely on AI (KGE models) while neglecting to integrate the invaluable insights that can be offered by humans who excel in conceptual analysis, critical thinking, and creativity. To exploit the merits of both AI and human, human-AI teaming (HAIT) was proposed, a convergence of human and AI capabilities into a collaborative system aiming to establish efficient and effective system across various domains, including data mining \cite{HAI-Comp}, decision making \cite{Game-1}, and text generating \cite{LLM_HAIT}. HAIT presents promising yet challenging opportunities that deserve further research and exploration.

In this work, we propose a novel HAIT system for LP, dubbed KG-HAIT, which combines humans' advantage to perform conceptual analyses of KGs and devise methodologies for extracting semantic information to drive logical reasoning, along with the promising efficacy in handling vast datasets demonstrated by AI. In KG-HAIT, humans design a graph dynamic programming (DP) based on their understanding of LP and KG and encapsulate these findings in human insightful feature (HIF) vectors. Then, HIF will be encoded into the training process of AI (a KGE model), resulting in better performance in LP that could be achieved by neither alone. The key contributions of our work include:

\begin{itemize}
    \item We propose KG-HAIT, a system with humans logical offering findings on the KG (human insightful feature, HIF, produced by a human-designed DP on KG) to enhance AI (the KGE model).
    
    KG-HAIT = HIF (human) + KGE model (AI).
    
    \item We evaluate KG-HAIT with three existing KGE models on three LP benchmarks, showcasing that human insights bring remarkable improvements with a notable reduction in the required number of training epochs.
    
    \item We demonstrate that HIF captures the semantic similarities between entities and can be adopted to distinguish between entity types.
\end{itemize}


\section{Related Works}


\subsection{Human-AI Teaming}

Humans and AI possess distinct ``cognitive frameworks'' \cite{HAI-Diff}, and recent endeavors to exploit this complementarity have proven to team humans and AI successfully in various domains. In data mining, \cite{HAI-Comp} asserted that human and AI's divergent perspectives in classification problems lead to misclassification of different kinds, and the combination of human and AI's prediction with extra confidence score improves accuracy. Meanwhile, the human and AI analyst teams in \cite{Stock-pred} outperform human analysts in collecting investment information and forecasting stock prices. For decision making, \cite{Game-1}, \cite{Game-2}, and \cite{Game-3} studied the HAIT mechanism's outcome, advancements, and social perception in the context of cooperative games with partially observable information. In the prevalent large language model (LLM) realm, the significant impact of human-in-the-loop (HILT) can never be understated. LLMs generate more reasonable content when their outputs are aligned with human feedback \cite{LLM-HITL}, \cite{LLM_HAIT}. Studies also report that HILT can promote AI's creativity \cite{HAI-Creativity}, which is hard to codify and quantify. One further instance is the well-known reinforcement learning from human feedback (RLHF) \cite{RLHF-Review}, an effective method of training models to act better in accordance with human preferences, which has found application in LLM, natural language processing (NLP), computer vision (CV), and game development.


\subsection{Dynamic Programming}

Dynamic Programming (DP) has been extensively studied \cite{DP}, \cite{DP-Apps-in-MS}, \cite{CR-DP} in mathematics and computer science, which models intricate problems into simplified states and solves through systematic breaking down of complexities into easier subproblems. States of DP are thoughtfully crafted \cite{ToDP} to extract desired information from complicated mathematical models, and its heterogeneous nature showcases DP's versatility demonstrating effectives in optimization \cite{DP-Optimization-1}, \cite{DP-Optimization-2}, counting \cite{DP-Counting}, scheduling \cite{DP-Scheduling-1}, \cite{DP-Scheduling-2} and more.

While research has revealed an intricate theoretic connection between DP and graph neural network (GNN) \cite{GNN-DPer}, in practice, DP properties cannot be found in trained GNN due to the limitation of training algorithms. Like many other AI models, GNN lacks interpretable and mathematical foundations. In contrast, DP has a relatively stronger theoretical basis and interpretability yet currently relies on a manual design by human experts. In fact, efforts have emerged to integrate DP and GNN for finding approximate solutions to NP problems \cite{DP-GNN-1}, \cite{DP-GNN-2}, representing the fusion of AI models and deterministic theoretic computing problems solving (AI helps humans). Conversely, AI models deserve better accuracy and faster convergence rate entailed by humans' established theories, and this motivates us to design a ``human helps AI'' paradigm backing up the link prediction model by the power of DP.




\subsection{Link Prediction Models} \label{subsec:KGE_Intro}

The past few years have seen the development of new Knowledge Graph Embedding (KGE) models of all kinds, and they can be categorized into three main types based on the architecture \cite{KGE_Summary}, which are geometric models (e.g. TransE \cite{TransE}, TransH \cite{TransH}, RotatE \cite{RotatE}), tensor factorization models (e.g. DistMult \cite{DistMult}, TuckER \cite{TuckER}), deep learning models (e.g. ConvE \cite{ConvE}, KBGAT \cite{KBGAT}, MoCoSA \cite{MoCoSA}).

KGE models ecnode each head entity $h$, relation $r$, and tail entity $t$ with low-dimensional embedding vectors $\mathbf h$, $\mathbf r$, and $\mathbf t$. Often case, they transform these embeddings $\mathbf h$, $\mathbf r$ and $\mathbf t$ into $f_h(\mathbf h)$, $f_r(\mathbf r)$ and $f_t(\mathbf t)$, and assign a score $s\langle h, r, t \rangle = \delta(f_h(\mathbf h), f_r(\mathbf r), f_t(\mathbf t))$ to triple $\langle h, r, t \rangle$. Note that the mapping $f_h$, $f_r$, $f_t$, and the scoring function $\delta$ represent the architecture of the model and are learned during training.

In this study, we focus on geometric models for two primary reasons: (1) Their structural simplicity and intuitive comprehensibility are crucial for humans to understand their AI teammates effectively, fostering harmonious cooperation. (2) While training algorithms may fall short of fully revealing a model's structural design, human intervention can guide it toward the intended direction, thus enhancing performance.

\subsubsection{Geometric Models}

Geometric models regard relations as geometric operations within latent spaces. The scoring function of geometric models depends on the spatial transformation $\psi(f_h, f_r)$, and the distance function $\delta'(\psi, f_t)$.

TransE \cite{TransE}, as the first geometric KGE model with $\psi(f_h, f_r) = f_h + f_r = \mathbf h + \mathbf r$, enforces $\mathbf t \approx \mathbf h + \mathbf r$, and measures the difference between $\mathbf h + \mathbf r$ and $\mathbf t$ by taking the $L^p$ norm. Despite its simplicity, it succeeded in link prediction and inspired lots of variants. An intuitive understanding of TransE is that the resulting embedding is an attempt to pin the KG in the $\mathbb R^d$ space ($d$ is the dimensionality of the embedding) while making sure that $f_r(\mathbf r_1 \circ \mathbf r_2 \circ \cdots \circ \mathbf r_k) \approx \sum_{i=1}^k f_r(\mathbf r_i)$ and $f_h(\mathbf h) + f_r(\mathbf r) \approx f_t(\mathbf t)$.



More details about these geometric models are shown in Table \ref{tab:models}.

\begin{table}[ht]
    \centering
    \caption{Architecture of Several Geometric KGE Models}
    \begin{tabular}{|c|c|c|c|c|} \hline
        Model & $f_h(\mathbf h)$ & $f_r(\mathbf r)$ & $f_t(\mathbf t)$ & $\delta(f_h, f_r, f_t)$ \\ \hline
        TransE & $\mathbf h$ & $\mathbf r$ & $\mathbf t$ & $\Vert f_h + f_r - f_t \Vert_p$ \\
        TransH & $\mathbf h - \mathbf n_r^{\top} \mathbf h  \mathbf n_r$ & $\mathbf r$ & $\mathbf t - \mathbf n_r^{\top} \mathbf t \mathbf n_r$ & $\Vert f_h + f_r - f_t \Vert_p$ \\
        TransR & $\mathbf M_r \times \mathbf h$ & $\mathbf r$ & $\mathbf M_r \times \mathbf t$ & $\Vert f_h + f_r - f_t \Vert_p$ \\ \hline
    \end{tabular}
    \label{tab:models}
\end{table}

\begin{figure*}[!ht]
    \centering
    \includegraphics[width = \textwidth]{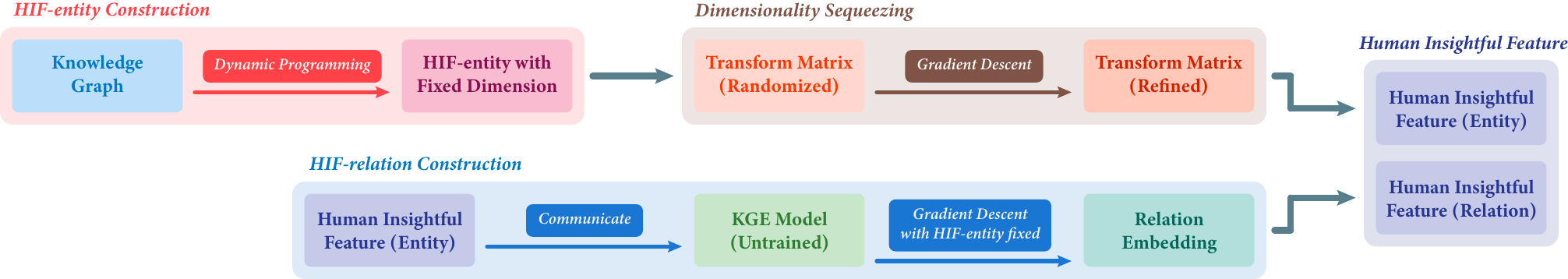}
    \caption{The workflow of how human insightful feature is obtained where it follows the order 1. HIF-entity construction, 2. dimensionality sequeezing, and 3. HIF-relation construction}
    \label{fig:struct}
    \vspace{-0.75\baselineskip}
\end{figure*}

\section{Methods}

Our proposed Human-AI Team system KG-HAIT consists of three parts illustrated in Fig.~\ref{fig:struct}. The first part extracts the human insightful feature vectors for entities (HIF-entity) from the given KG, and the second part finds an appropriate transform matrix to adjust the dimensionality of HIF-entity to any desired number. In the third part, the HIF-entity are used to find HIF-relation. Ultimately, the AI part (a KGE model) is trained based on both HIF-entity and HIF-relation.

\subsection{Notation}

Knowledge graph (KG) is a directed heterogeneous graph $G = (\mathcal E, \mathcal R, \mathcal T)$, where $\mathcal E$ denotes the set of nodes representing entities, $\mathcal R$, the set of labels representing relations (predicates), and $\mathcal T \subseteq \mathcal E \times \mathcal R \times \mathcal E$, the set of triples. Each triple $\langle h, r, t \rangle$ is ordered and signifies the existence of a directed edge labeled $r$ pointing from $h$ to $t$.

\subsection{HIF-entity Construction}

\begin{figure*}[!t]
    \centering
    \includegraphics[width = 0.8\textwidth]{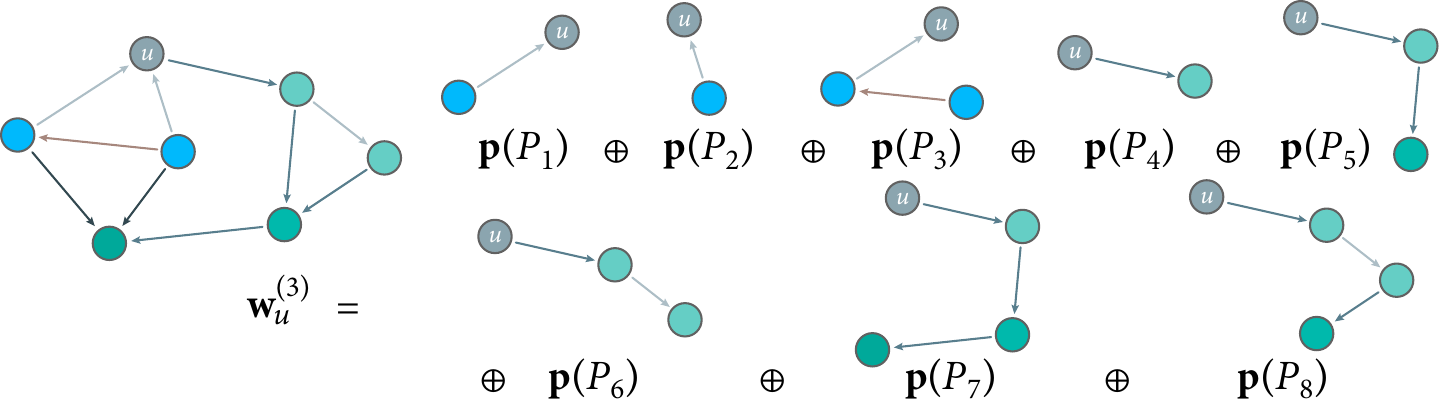}
    \caption{An example of how HIF-entity is calculated. The DP result after 3 iterations $\mathbf w^{(3)}_u$ is obtained by applying general addition on $\mathbf p(P_1)$, $\mathbf p(P_2)$, $\cdots$, $\mathbf p(P_8)$. Additionaly, $\mathbf p(P_k)$, $k \in [1, 8] \cap \mathbb Z$ is defined as the general product of all the entities and relations included in that path.}
    \label{fig:DP_example}
    \vspace{-0.75\baselineskip}
\end{figure*}

Fully human-designed dynamic programming (DP) is leveraged to collect and aggregate the graph structure surrounding each entity to produce HIF-entity. For simplicity, define the head, relation, and tail extraction function $\mathfrak h : \mathcal T \to \mathcal E$, $\mathfrak r : \mathcal T \to \mathcal R$, and $\mathfrak t : \mathcal T \to \mathcal E$ returning the respective head, relation, and tail for triple $p = \langle h, r, t \rangle$,
$$
\mathfrak h p = \mathfrak h \langle h, r, t \rangle = h
\quad
\mathfrak r p = \mathfrak r \langle h, r, t \rangle = r
\quad
\mathfrak t p = \mathfrak t \langle h, r, t \rangle = t
$$
also denote $\mathcal M^{in}_u$ and $\mathcal M^{out}_u$ as the set of in-coming and out-going neighboring triples of entity $u$, namely,
$$
\mathcal M^{in}_u \triangleq \{ r \in \mathcal T \mid \mathfrak t r = u \}
\quad
\mathcal M^{out}_u \triangleq \{ r \in \mathcal T \mid \mathfrak h r = u \}
$$

It would be comfortable to communicate with AI by concentrating the graph information around entity $u$ into a $d$-dimensional vector $\mathbf w_u^{(t)}$ being the result of DP (HIF-entity) for entity $u$ after $t$ iterations, and $u$ would gradually interact with its neighbors as interation continues.

Firstly, when $t = 1$, $\mathbf w_u^{(1)}$ depends on entity $u$ only, which states, (followed by the actual adopted function)
$$
\mathbf w_{u}^{(1)} \triangleq \mathbf e(u)
=
\sum_{p \in \mathcal M^{out}_u} \mathbf o(\mathfrak r p)
-
\sum_{p \in \mathcal M^{in}_u} \mathbf o(\mathfrak r p)
$$
where $\mathbf e(u) : \mathcal E \to \mathbb R^d$ is the entity identity function and $\mathbf o(\mathfrak r p)$ is the one-hot encoding for $\mathfrak r p$.

For larger $t$, $\mathbf w_u^{(t)}$ would extract more information from its neighborhood. The recursive DP transition is designed as
$$ \begin{aligned}
\mathbf w_u^{(t+1)} = \left(
    \bigoplus_{p \in \mathcal M^{in}_u} \mathbf w_{\mathfrak h p}^{(t)} \otimes \textbf v(p)
\right) \oplus \left(
    \bigoplus_{p \in \mathcal M^{out}_u} \textbf v(p) \otimes \mathbf w_{\mathfrak t p}^{(t)}
\right)
\end{aligned} $$
where $\oplus$ and $\otimes$ are general addition and multiplication which is determined by humans, and $\mathbf v(p) : \mathcal T \to \mathbb R^d$ presenting the vectorial weight of a triple. Detailed process can be found in Algorithm 1.

\begin{figure}[H]
\vspace{-1.5\baselineskip}
\begin{algorithm}[H]
\footnotesize
\caption{Construct HIF-entity}
\begin{algorithmic}[1]
\Require
\begin{itemize}
    \item[]
    \item $G = (\mathcal E, \mathcal R, \mathcal T)$, the KG;
    \item $\mathbf e(u), \mathbf v(p)$, the entity identity and triple weight function;
    \item $\otimes$, $\oplus$, the general multiplication and addition operator;
    \item $T$ the iteration upper bound
\end{itemize}

\Ensure
\begin{itemize}
    \item[]
    \item $\mathbf w_{u}^{(T)}$, the HIF-entity for each entity $u$.
    \item[]
\end{itemize}

\For{$u \in \mathcal E$}
    \State $\mathbf w_u^{(1)} \gets \mathbf e(u)$
\EndFor
\For{$t \gets 2 \cdots T$}
    \For{$u \in \mathcal E$}
        \State $\mathbf w^{in} \gets \mathbf e(u)$, $\mathbf w^{out} \gets \mathbf e(u)$
        \For{$p \in \mathcal M^{in}_u$}
            \State $\mathbf w^{in} \gets \mathbf w^{in} \oplus \left( \mathbf w^{(t-1)}_{\mathfrak h p} \otimes \mathbf v(p) \right)$
        \EndFor
        \For{$p \in \mathcal M^{out}_u$}
            \State $\mathbf w^{out} \gets \mathbf w^{out} \oplus \left( \mathbf v(p) \otimes \mathbf w^{(t-1)}_{\mathfrak t p} \right)$
        \EndFor
        \State $\mathbf w_u^{(t)} = \mathbf w^{in} \oplus \mathbf w^{out}$
    \EndFor
\EndFor
\end{algorithmic}
\end{algorithm}

\vspace{-1.5\baselineskip}
\end{figure}

It can be shown by mathematical induction that $\mathbf w_u^{(t)}$ contains information about the $t$-neighborhood $\mathcal N^{(t)}_u$ of $u$, for
$$
\mathcal N^{(t)}_u \triangleq \{ v \in \mathcal E \mid d(u, v) \le t \}
$$
where $d(u, v)$ be the distance between $u$ and $v$.

Also, $\mathbf w_{u}^{(t)}$ ensembles information of every single path starting from $u$ or ending at $u$, even though the quantity of such paths could be exponential. Let $\mathcal P^{in,(t)}_u$ and $\mathcal P^{out,(t)}_u$ be the set of paths ending or starting at $u$ with a length no longer than $t$, and $\mathbf p(P)$ is called a path feature vector (PFV) for path $P = \{ p_1, p_2 \cdots, p_s \}$ which operate the general multiplication for all entities and relations along the path,
$$ \mathbf p(P) \triangleq \mathbf e(\mathfrak h p_1) \otimes \bigotimes_{p_i \in P} \mathbf v(p_i) \otimes \mathbf e(\mathfrak t p_i) $$
then $\mathbf w^{(t)}_u$ aggregate all PFV with the general addition,
$$
\mathbf w^{(t)}_u =
\left( \bigoplus_{P \in \mathcal P^{in, (t)}_u} \mathbf p(P) \right)
\oplus
\left( \bigoplus_{P \in \mathcal P^{out, (t)}_u} \mathbf p(P) \right)
$$
An intuitive example can be found in Fig.~\ref{fig:DP_example}


In this study, the dimensionality of $\mathbf w^{(t)}_u$ is $d = | \mathcal R |$ where each dimension corresponds to a unique relation, denoting as the intensity of that relation. e.g. $\mathbf w_{u, r}^{(t)}$, defines the magnitude of effect relation $r$ have on entity $u$, and such intensity should somehow depend on the presence of $r$ in $\mathcal N^{(t)}_u$. Human intuition suggest adopting $\mathbf v(p) = \alpha$, which is a decaying rate constant for discouraging contributions from distant entities. Additionally, $\otimes$ is regarded as scalar multiplication for vector and $\oplus$ as the element-wise maximum. Furthermore, to distinguish between in-coming relations and out-going relations, the contribution from out-going relations are positive while the in-coming ones are negative. Formally,

$$
\mathbf w_u^{(t)} =
\max_{p \in \mathcal M^{out}_u} \left\{ \alpha \circ \mathbf w_{\mathfrak t p}^{(t)} \right\}
-
\max_{p \in \mathcal M^{in}_u} \left\{ \alpha \circ \mathbf w_{\mathfrak h p}^{(t)} \right\}
$$
$$ (\alpha \circ \mathbf b)_k = \alpha \mathbf b_k \quad \max\{ \mathbf a, \mathbf b \}_k = \max\{ \mathbf a_k, \mathbf b_k \} $$

When bounding the maximal iteration by $T$, the final outcome of DP $\mathbf w_{1 \cdots |\mathcal E|}^{(T)}$ represents the HIF-entity with fixed dimensionality. We posit, and intend to empirically substantiate that HIF-entity $\mathbf e_{1 \cdots |\mathcal E|} = \mathbf w_{1 \cdots |\mathcal E|}^{(T)}$ could extract the semantic meaning inherent in the KG. This assertion is underpinned by the understanding that $\mathbf e_{u}$, obtained through applying DP on the local subgraph surrounding entity $u$, encapsulates local subgraph feature such as ensemble information about paths around an entity, closeness and intensity with respect to each relation, etc. Importantly, a properly constructed KG implies that entities sharing similar semantic meaning tend to lead exhibit similar local subgraph feature, and vice versa. Consequently, HIF-entity which captures local subgraph features inherently encodes semnatic similarities.

\subsection{Dimensionality Sequeezing}

The obtained $\mathbf w_u^{(T)}$ has a fixed dimensionality $d = |\mathcal R|$, and cannot be directly used as an entity embedding as it might differ from the KGE model's dimensionality of entity $d_e$. Thus, we need to find a transformation matrix $\mathbf M \in \mathbb R^{d_e \times |\mathcal R|}$ s.t.
$$
\mathbf e_u = \mathbf M \times \mathbf w_u^{(T)}
= \sum_{k=1}^{|\mathcal R|} \mathbf w_{u, k}^{(T)} \mathbf c_k
$$
where $\mathbf e_u$ denotes the entity embedding to be used in training for entity $u$, and $\mathbf c_k$, the $k$-th column vector of $\mathbf M$.

Randomly generated $\mathbf c_{1 \cdots |\mathcal R|}$ might incorrectly map different vectors $\mathbf v_1$, $\mathbf v_2$ into similar ones with small $\cos \langle \mathbf M \mathbf v_1, \mathbf M \mathbf v_2 \rangle$. To preserve the property of HIF-entity to the greatest extent, the following criterion should be satisfied
$$ \cos\langle \mathbf M \mathbf v_1, \mathbf M \mathbf v_2 \rangle \approx \cos\langle \mathbf v_1, \mathbf v_2 \rangle $$

Human insights suggest that the criterion could be achieved when $\mathbf c_{1 \cdots |\mathcal R|}$ are mostly pairwise orthogonal, i.e., the cosine similarity between these column vectors should be close to 0. Simple gradient descent method with loss $\mathcal L_{mcs}$ can minimize the maximal absolute cosine similarity between these column vectors making dimensionality sequeezing reliable:
$$ \mathcal L_{mcs} = \max_{\mathbf c_i, \mathbf c_j}\{ |\cos\langle \mathbf c_i, \mathbf c_j \rangle| \} = \max_{\mathbf c_i, \mathbf c_j}\left\{ \frac{|\mathbf c_i^{\top} \mathbf c_j|}{\Vert \mathbf c_i \Vert_2 \cdot \Vert \mathbf c_j \Vert_2} \right\} $$

\subsection{HIF-relation Construction}


The construction of HIF-relation poses challenge because 1) KGs typically contain fewer distinct relations compared to entities and 2) a certain relation appears multiple times throughout the KG making it clueless to extract a local subgraph feature specific each relation. However, we can take advantage of HIF-entity and the model itself, which is to say, HIF-relation $\mathbf r_{1 \cdots |\mathcal R|}$ is constructed by intializing a KGE model $\mathcal M$ with fixed entity embedding offered by HIF-entity $\mathbf E = \mathbf e_{1 \cdots |\mathcal E|}$, training it for multiple epoches, and saving its relation embedding $\mathbf R$.

\setcounter{table}{2}

\begin{table*}[t!]
\centering
\caption{Result of applying HIF on TransE, TransH, TransR, and testing on FB15k-237, WN18RR and LastFM-9}
\begin{tabular}{|c|ccccc|ccccc|ccccc|} \hline
\multirow{2}{*}{\diagbox{Model}{Dataset}} & \multicolumn{5}{c|}{FB15k237} & \multicolumn{5}{c|}{WN18RR} & \multicolumn{5}{c|}{LastFM-9} \\ \cline{2-16}
                & MR            & MRR           & H@1           & H@3           & H@10
                & MR            & MRR           & H@1           & H@3           & H@10
                & MR            & MRR           & H@1           & H@3           & H@10
\\ \hline
TransE w/o HIF  & 186           & .289          & .192          & .326          & .478
                & 3787          & .176          & .007          & .323          & .450
                & 554           & .603          & .449          & .732          & .805          \\
TransE w/ HIF   & \textbf{135}  & \textbf{.329} & \textbf{.232} & \textbf{.368} & \textbf{.519}
                & \textbf{2271} & \textbf{.202} & \textbf{.037} & \textbf{.331} & \textbf{.485}
                & \textbf{307}  & \textbf{.703} & \textbf{.636} & \textbf{.739} & \textbf{.836}
\\ \hline
TransH w/o HIF  & 180           & .291          & .191          & .325          & .485
                & 3809          & .182          & .011          & .333          & .457
                & 667           & .592          & .447          & .716          & .811

\\
TransH w/ HIF   & \textbf{134}  & \textbf{.332} & \textbf{.237} & \textbf{.369} & \textbf{.520}
                & \textbf{1988} & \textbf{.223} & \textbf{.044} & \textbf{.374} & \textbf{.503}
                & \textbf{225}  & \textbf{.752} & \textbf{.698} & \textbf{.784} & \textbf{.855}

\\ \hline
TransR w/o HIF  & 182           & .306          & .209          & .344          & .497
                & 3751          & .184          & .011          & .336          & .456
                & 586           & .652          & .522          & .761          & .848
\\
TransR w/ HIF   & \textbf{119}  & \textbf{.357} & \textbf{.258} & \textbf{.400} & \textbf{.552}
                & \textbf{2242} & \textbf{.206} & \textbf{.037} & \textbf{.339} & \textbf{.486}
                & \textbf{268}  & \textbf{.756} & \textbf{.701} & \textbf{.788} & \textbf{.870}
\\ \hline
\end{tabular}
\label{tab:LP_res}

\vspace{-0.75\baselineskip}
\end{table*}

Let $\mathbf E = \mathbf e_{1 \cdots |\mathcal E|}$ be the weight matrix for entity embeddings, $\mathcal M(\mathbf E, \mathbf R)$ be a KGE model with entity embedding $\mathbf E$ and relation embedding $\mathbf R$, and $\mathcal L(\mathcal M, \mathcal T)$ be the loss defined for model $\mathcal M$ on dataset $\mathcal T$. Then, HIF-relation is obtained by
$$ \mathbf r_{1 \cdots |\mathcal R|} = \arg\min_{\mathbf R \in \mathbb R^{|\mathcal R| \times d_r}} \left\{ \mathcal L(\mathcal M \left( \mathbf E, \mathbf R), \mathcal T_{train} \right) \right\} $$
this can be numerically obatined by gradient descent but with entity embedding $\mathbf E$ (which is HIF-entity $\mathbf e_{1 \cdots |\mathcal E|}$ in this case) provided and fixed. In such a way, $\mathbf r_{1 \cdots |\mathcal R|}$ inherits the human insights encoded in $\mathbf e_{1 \cdots |\mathcal E|}$, thereby facilitating smooth training of the AI teammate (the KGE model).

\section{Experiments}

\subsection{Datasets}

We choose FB15k-237, WN18RR, and LastFM-9 as testing datasets to show the efficacy of human insights. FB15k-237 is extracted by \cite{TransE} from FreeBase, and improved by \cite{FB15k-237} to avoid the test leakage issue, being known as a challenging dataset. WN18RR was built by \cite{ConvE} based on WN18 with a similar test leakage issue fixed. LastFM-9 is constructed by \cite{KGAT} from a music listening dataset collected from last.fm, an online music platform. And we further randomly retain $75\%$ triples from LastFM-9 as $\mathcal T_{train}$ and blindly split the other $25\%$ triples into $\mathcal T_{valid}$ and $\mathcal T_{test}$.

Technical details of these datasets can be found in Table \ref{tab:dataset}.

\setcounter{table}{1}

\begin{table}[H]
    \vspace{-\baselineskip}
    \centering
    \caption{number of entities, relations, and triples in training, validation, and testing data of each dataset}
    \begin{tabular}{|c|ccccc|} \hline
        Dataset Name & $|\mathcal E|$ & $|\mathcal R|$ & $|\mathcal T_{train}|$ & $|\mathcal T_{valid}|$ & $|\mathcal T_{test}|$ \\ \hline
        FB15k-237   & 14541     & 237   & 272115    & 17536     & 20466     \\ \hline
        WN18RR      & 40943     & 11    & 86835     & 3034      & 3134      \\ \hline
        LastFM-9    & 106389    & 9     & 272011    & 46287     & 46270     \\ \hline
    \end{tabular}
    \label{tab:dataset}
\end{table}

\setcounter{table}{3}



\subsection{Implementation Details}

All models were implemented in PyTorch and trained on GPU (NVIDIA GeForce RTX 3090, Memory 24GB). The training was performed using mini-batch (batch size equals 2000) gradient descent optimizer Adam. Unless otherwise stated, All models have an entity embedding dimension $d_e = 100$ and relation one $d_r = 100$. We evaluate models with 
hyperparameters selected by grid search with: $\delta \in \{ \Vert \cdot \Vert_1, \Vert \cdot \Vert_2 \}$, $T \in \{2, 4, 6, 8, 12\}$, $\text{learning rate} \in \{ .002, .0002, .0002 \}$. The training would cease depending on evaluating the result of the validation set to prevent overfitting.

Note that TransR requires a pre-trained TransE model. To assess the outcome of human intervention on TransR, we let TransR w/o HIF inherit and TransR w/ HIF inherit from TransE w/o HIF and TransE w/ HIF, respectively.

\subsection{Results}

\subsubsection{Link Prediction}

To evaluate the efficacy of human teammate's insights, we evaluated HIF on TransE, TransH, and TransR and compared them to the original ones (i.e., models without HIF). As suggested by \cite{TransE}, we follow the filter protocol, and report metrics such as mean rank (MR, lower is better), mean reciprocal rank (MRR, higher is better), and hits at $k$ (H@$k$, higher is better) for $k \in \{ 1, 3, 10 \}$. The detailed results are listed in Table \ref{tab:LP_res}, which show that human insights (HIF) bring remarkable improvements to all models in all datasets and all metrics.

The greatest improvements are found in the metric MR, which has an average of $42.8\%$ decrease. In addition to that, HIF excels in boosting the H@1 score, bringing an about 4 times advancement in H@1 in WN18RR, an averaged 44\% promotion in LastFM-9, and a more than 20\% increase in FB15k237. HIF benefits these models in other metrics (MRR, H@3, H@10) as well. For FB15k237, HIF brings about a considerable 0.3-0.5 absolute increment in these metrics. Meanwhile, in WN18RR and LastFM-9, HIF also brought notable improvements to these models in these metrics.

It is noteworthy that TransR w/ HIF surpasses the performance of TransR w/o HIF, highlighting the inheritance of human insights. In other words, the performance gain obtained from HIF can be propagated across models.

\subsubsection{Semantic Similarity}

Recall that HIF is derived from the KG with humans understanding of graph structures. We empirically show that these insights indeed capture the semantic similarities of entities in the KG. Refer to Fig.~\ref{fig:KG} for an illustration where entities of the same type exhibit similar subgraph structures (neighborhood). For example, entity \texttt{Claude Shannon} and \texttt{Richard Feynman} are both human beings, so they both have the outgoing \texttt{BornIn} and \texttt{StudyIn} relation, and both are next to a university (\texttt{MIT}) and a country (\texttt{USA}). Therefore, their corresponding HIF-entity vectors would have high cosine similarities, indicating they might share the same type. Thus, entities with closely aligned HIF vectors (high cosine similarity) share analogous semantic meanings, implying similar types.

\begin{table}[ht]
    \scriptsize
    \centering
    \caption{List of selected entities}
    \begin{tabular}{|c|c|} \hline
        Country / Region ($\mathbf p_i$) & Educational Institution ($\mathbf q_i$) \\ \hline
        Dominican Republic & Case Western Reserve University \\
        Japan & Taras Shevchenko National University of Kyiv \\
        United States of America & University of Sussex \\
        Bosnia and Herzegovina & California State University, Northridge \\
        Bangladesh & Ohio University \\ \hline
    \end{tabular}
    \label{tab:selected}
    \vspace{-1.5\baselineskip}
\end{table}

\begin{figure}[ht]
    \vspace{-0.5\baselineskip}
    \centering
    \input{fig2.pgf}
    \vspace{-1.5\baselineskip}
    \caption{The confusion matrices of cosine similarity between HIF-entity for 2 types of entities representing country/region and educational institution}
    \label{fig:confusion_matrices}
\end{figure}
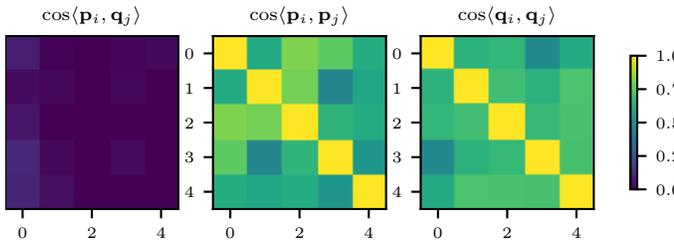

To assess HIF's classification capability, we randomly selected two types of entities in FB15k-237 — county/region and educational institution. Then, we chose 5 random entities of each type and calculated the inner product of each pair. For simplicity, HIF-entity vectors from country/region are denoted as $\mathbf p_i$ and that of the educational institution as $\mathbf q_i$ where $ i \in [1, 5] \cap \mathbb Z$ (Table \ref{tab:selected}). The result of entity similarity is displayed in Fig.~\ref{fig:confusion_matrices} as confusion matrices. The average cosine similarity across ``country/region'' and ``educational institution'' is $0.656\%$. Specifically, within the "country/region" category, the average similarity is $71.36\%$;  within the "educational institution" category, it is $72.30\%$. The extraction of semantic meaning with the aid of humans might be key to the improvement of KGE model's performance in link prediction.

\subsubsection{Convergence Rate}

\begin{figure}[ht]
    \vspace{-\baselineskip}
    \centering
    \input{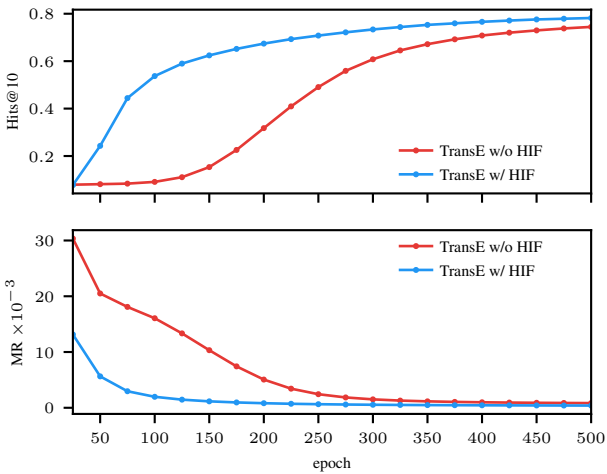}
    \caption{H@10 and MR for every 25 epochs (TransE w/o HIF and TransE w/ HIF on LastFM-9)}
    \label{fig:curves}
\end{figure}

The outcomes of TransE w/o HIF and TransE w/ HIF on H@10 and MR over 500 epochs are illustrated in Fig.~\ref{fig:curves} for LastFM-9, featuring an embedding dimension of 50 and a learning rate of $5 \times 10^{-4}$.

Under the metric H@10, TransE w/ HIF grows fast in the first 100 epochs and tends to converge after epoch 200. In contrast, TransE continues to proliferate and shows no indications of convergence even up to epoch 400. Concerning the MR metric, TransE w/ HIF begins to converge around epoch 150, whereas TransE shows no signs of convergence until epoch 300. It demonstrates that KG-HAIT is capable of achieving effective training outcomes while significantly reducing the number of required epochs.

\section{Conclusion}

In this study, we introduce an innovative human-AI team method dubbed KG-HAIT aimed at addressing the long-overlooked issue of lacking human insights in link prediction (LP). In KG-HAIT, humans reveal the insightful feature of the knowledge graph to AI via human insightful feature (HIF), which harnesses dynamic programming (DP) for its effectiveness in aggregating the graph characteristics. In experiments, KG-HAIT demonstrates a significant enhancement in the performance of various translational KGE models across LP benchmarks in all evaluation metrics. This success can be attributed to humans' capability to capture the semantic meaning of entities in the KG, which also enables HIF to differentiate between diverse entity types. Furthermore, results show that KG-HAIT accelerates model convergence rate, facilitating smoother and more efficient training.

Looking ahead, our future work involves extending the human-AI team mechanism to more models, specifically focusing on more sophisticated ones. Furthermore, we look forward to delving deeper into human insightful feature vectors and finding the precise underlying reason why they boost the performance of KGE models. Ultimately, we aspire to investigate the HAIT mechanism further and develop a collaborative system wherein humans and AI can cooperate more closely on KG.


\addtolength{\textheight}{-9cm}

\bibliography{citation}

\end{document}

%% file: fig2.pgf
\begingroup%
\makeatletter%
\begin{pgfpicture}%
\pgfpathrectangle{\pgfpointorigin}{\pgfqpoint{3.850000in}{1.400000in}}%
\pgfusepath{use as bounding box, clip}%
\begin{pgfscope}%
\pgfsetbuttcap%
\pgfsetmiterjoin%
\definecolor{currentfill}{rgb}{1.000000,1.000000,1.000000}%
\pgfsetfillcolor{currentfill}%
\pgfsetlinewidth{0.000000pt}%
\definecolor{currentstroke}{rgb}{1.000000,1.000000,1.000000}%
\pgfsetstrokecolor{currentstroke}%
\pgfsetdash{}{0pt}%
\pgfpathmoveto{\pgfqpoint{0.000000in}{0.000000in}}%
\pgfpathlineto{\pgfqpoint{3.850000in}{0.000000in}}%
\pgfpathlineto{\pgfqpoint{3.850000in}{1.400000in}}%
\pgfpathlineto{\pgfqpoint{0.000000in}{1.400000in}}%
\pgfpathlineto{\pgfqpoint{0.000000in}{0.000000in}}%
\pgfpathclose%
\pgfusepath{fill}%
\end{pgfscope}%
\begin{pgfscope}%
\pgfsetbuttcap%
\pgfsetmiterjoin%
\definecolor{currentfill}{rgb}{1.000000,1.000000,1.000000}%
\pgfsetfillcolor{currentfill}%
\pgfsetlinewidth{0.000000pt}%
\definecolor{currentstroke}{rgb}{0.000000,0.000000,0.000000}%
\pgfsetstrokecolor{currentstroke}%
\pgfsetstrokeopacity{0.000000}%
\pgfsetdash{}{0pt}%
\pgfpathmoveto{\pgfqpoint{0.000000in}{0.247059in}}%
\pgfpathlineto{\pgfqpoint{0.905882in}{0.247059in}}%
\pgfpathlineto{\pgfqpoint{0.905882in}{1.152941in}}%
\pgfpathlineto{\pgfqpoint{0.000000in}{1.152941in}}%
\pgfpathlineto{\pgfqpoint{0.000000in}{0.247059in}}%
\pgfpathclose%
\pgfusepath{fill}%
\end{pgfscope}%
\begin{pgfscope}%
\pgfpathrectangle{\pgfqpoint{0.000000in}{0.247059in}}{\pgfqpoint{0.905882in}{0.905882in}}%
\pgfusepath{clip}%
\pgfsys@transformshift{0.000000in}{0.247059in}%
\pgftext[left,bottom]{\includegraphics[interpolate=true,width=0.910000in,height=0.910000in]{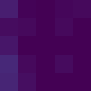}}%
\end{pgfscope}%
\begin{pgfscope}%
\pgfsetbuttcap%
\pgfsetroundjoin%
\definecolor{currentfill}{rgb}{0.000000,0.000000,0.000000}%
\pgfsetfillcolor{currentfill}%
\pgfsetlinewidth{0.803000pt}%
\definecolor{currentstroke}{rgb}{0.000000,0.000000,0.000000}%
\pgfsetstrokecolor{currentstroke}%
\pgfsetdash{}{0pt}%
\pgfsys@defobject{currentmarker}{\pgfqpoint{0.000000in}{-0.048611in}}{\pgfqpoint{0.000000in}{0.000000in}}{%
\pgfpathmoveto{\pgfqpoint{0.000000in}{0.000000in}}%
\pgfpathlineto{\pgfqpoint{0.000000in}{-0.048611in}}%
\pgfusepath{stroke,fill}%
}%
\begin{pgfscope}%
\pgfsys@transformshift{0.090588in}{0.247059in}%
\pgfsys@useobject{currentmarker}{}%
\end{pgfscope}%
\end{pgfscope}%
\begin{pgfscope}%
\definecolor{textcolor}{rgb}{0.000000,0.000000,0.000000}%
\pgfsetstrokecolor{textcolor}%
\pgfsetfillcolor{textcolor}%
\pgftext[x=0.090588in,y=0.149837in,,top]{\color{textcolor}\rmfamily\fontsize{6.000000}{7.200000}\selectfont \(\displaystyle {0}\)}%
\end{pgfscope}%
\begin{pgfscope}%
\pgfsetbuttcap%
\pgfsetroundjoin%
\definecolor{currentfill}{rgb}{0.000000,0.000000,0.000000}%
\pgfsetfillcolor{currentfill}%
\pgfsetlinewidth{0.803000pt}%
\definecolor{currentstroke}{rgb}{0.000000,0.000000,0.000000}%
\pgfsetstrokecolor{currentstroke}%
\pgfsetdash{}{0pt}%
\pgfsys@defobject{currentmarker}{\pgfqpoint{0.000000in}{-0.048611in}}{\pgfqpoint{0.000000in}{0.000000in}}{%
\pgfpathmoveto{\pgfqpoint{0.000000in}{0.000000in}}%
\pgfpathlineto{\pgfqpoint{0.000000in}{-0.048611in}}%
\pgfusepath{stroke,fill}%
}%
\begin{pgfscope}%
\pgfsys@transformshift{0.452941in}{0.247059in}%
\pgfsys@useobject{currentmarker}{}%
\end{pgfscope}%
\end{pgfscope}%
\begin{pgfscope}%
\definecolor{textcolor}{rgb}{0.000000,0.000000,0.000000}%
\pgfsetstrokecolor{textcolor}%
\pgfsetfillcolor{textcolor}%
\pgftext[x=0.452941in,y=0.149837in,,top]{\color{textcolor}\rmfamily\fontsize{6.000000}{7.200000}\selectfont \(\displaystyle {2}\)}%
\end{pgfscope}%
\begin{pgfscope}%
\pgfsetbuttcap%
\pgfsetroundjoin%
\definecolor{currentfill}{rgb}{0.000000,0.000000,0.000000}%
\pgfsetfillcolor{currentfill}%
\pgfsetlinewidth{0.803000pt}%
\definecolor{currentstroke}{rgb}{0.000000,0.000000,0.000000}%
\pgfsetstrokecolor{currentstroke}%
\pgfsetdash{}{0pt}%
\pgfsys@defobject{currentmarker}{\pgfqpoint{0.000000in}{-0.048611in}}{\pgfqpoint{0.000000in}{0.000000in}}{%
\pgfpathmoveto{\pgfqpoint{0.000000in}{0.000000in}}%
\pgfpathlineto{\pgfqpoint{0.000000in}{-0.048611in}}%
\pgfusepath{stroke,fill}%
}%
\begin{pgfscope}%
\pgfsys@transformshift{0.815294in}{0.247059in}%
\pgfsys@useobject{currentmarker}{}%
\end{pgfscope}%
\end{pgfscope}%
\begin{pgfscope}%
\definecolor{textcolor}{rgb}{0.000000,0.000000,0.000000}%
\pgfsetstrokecolor{textcolor}%
\pgfsetfillcolor{textcolor}%
\pgftext[x=0.815294in,y=0.149837in,,top]{\color{textcolor}\rmfamily\fontsize{6.000000}{7.200000}\selectfont \(\displaystyle {4}\)}%
\end{pgfscope}%
\begin{pgfscope}%
\pgfsetbuttcap%
\pgfsetroundjoin%
\definecolor{currentfill}{rgb}{0.000000,0.000000,0.000000}%
\pgfsetfillcolor{currentfill}%
\pgfsetlinewidth{0.803000pt}%
\definecolor{currentstroke}{rgb}{0.000000,0.000000,0.000000}%
\pgfsetstrokecolor{currentstroke}%
\pgfsetdash{}{0pt}%
\pgfsys@defobject{currentmarker}{\pgfqpoint{-0.048611in}{0.000000in}}{\pgfqpoint{-0.000000in}{0.000000in}}{%
\pgfpathmoveto{\pgfqpoint{-0.000000in}{0.000000in}}%
\pgfpathlineto{\pgfqpoint{-0.048611in}{0.000000in}}%
\pgfusepath{stroke,fill}%
}%
\begin{pgfscope}%
\pgfsys@transformshift{0.000000in}{1.062353in}%
\pgfsys@useobject{currentmarker}{}%
\end{pgfscope}%
\end{pgfscope}%
\begin{pgfscope}%
\definecolor{textcolor}{rgb}{0.000000,0.000000,0.000000}%
\pgfsetstrokecolor{textcolor}%
\pgfsetfillcolor{textcolor}%
\pgftext[x=-0.148147in, y=1.033418in, left, base]{\color{textcolor}\rmfamily\fontsize{6.000000}{7.200000}\selectfont \(\displaystyle {0}\)}%
\end{pgfscope}%
\begin{pgfscope}%
\pgfsetbuttcap%
\pgfsetroundjoin%
\definecolor{currentfill}{rgb}{0.000000,0.000000,0.000000}%
\pgfsetfillcolor{currentfill}%
\pgfsetlinewidth{0.803000pt}%
\definecolor{currentstroke}{rgb}{0.000000,0.000000,0.000000}%
\pgfsetstrokecolor{currentstroke}%
\pgfsetdash{}{0pt}%
\pgfsys@defobject{currentmarker}{\pgfqpoint{-0.048611in}{0.000000in}}{\pgfqpoint{-0.000000in}{0.000000in}}{%
\pgfpathmoveto{\pgfqpoint{-0.000000in}{0.000000in}}%
\pgfpathlineto{\pgfqpoint{-0.048611in}{0.000000in}}%
\pgfusepath{stroke,fill}%
}%
\begin{pgfscope}%
\pgfsys@transformshift{0.000000in}{0.881176in}%
\pgfsys@useobject{currentmarker}{}%
\end{pgfscope}%
\end{pgfscope}%
\begin{pgfscope}%
\definecolor{textcolor}{rgb}{0.000000,0.000000,0.000000}%
\pgfsetstrokecolor{textcolor}%
\pgfsetfillcolor{textcolor}%
\pgftext[x=-0.148147in, y=0.852241in, left, base]{\color{textcolor}\rmfamily\fontsize{6.000000}{7.200000}\selectfont \(\displaystyle {1}\)}%
\end{pgfscope}%
\begin{pgfscope}%
\pgfsetbuttcap%
\pgfsetroundjoin%
\definecolor{currentfill}{rgb}{0.000000,0.000000,0.000000}%
\pgfsetfillcolor{currentfill}%
\pgfsetlinewidth{0.803000pt}%
\definecolor{currentstroke}{rgb}{0.000000,0.000000,0.000000}%
\pgfsetstrokecolor{currentstroke}%
\pgfsetdash{}{0pt}%
\pgfsys@defobject{currentmarker}{\pgfqpoint{-0.048611in}{0.000000in}}{\pgfqpoint{-0.000000in}{0.000000in}}{%
\pgfpathmoveto{\pgfqpoint{-0.000000in}{0.000000in}}%
\pgfpathlineto{\pgfqpoint{-0.048611in}{0.000000in}}%
\pgfusepath{stroke,fill}%
}%
\begin{pgfscope}%
\pgfsys@transformshift{0.000000in}{0.700000in}%
\pgfsys@useobject{currentmarker}{}%
\end{pgfscope}%
\end{pgfscope}%
\begin{pgfscope}%
\definecolor{textcolor}{rgb}{0.000000,0.000000,0.000000}%
\pgfsetstrokecolor{textcolor}%
\pgfsetfillcolor{textcolor}%
\pgftext[x=-0.148147in, y=0.671065in, left, base]{\color{textcolor}\rmfamily\fontsize{6.000000}{7.200000}\selectfont \(\displaystyle {2}\)}%
\end{pgfscope}%
\begin{pgfscope}%
\pgfsetbuttcap%
\pgfsetroundjoin%
\definecolor{currentfill}{rgb}{0.000000,0.000000,0.000000}%
\pgfsetfillcolor{currentfill}%
\pgfsetlinewidth{0.803000pt}%
\definecolor{currentstroke}{rgb}{0.000000,0.000000,0.000000}%
\pgfsetstrokecolor{currentstroke}%
\pgfsetdash{}{0pt}%
\pgfsys@defobject{currentmarker}{\pgfqpoint{-0.048611in}{0.000000in}}{\pgfqpoint{-0.000000in}{0.000000in}}{%
\pgfpathmoveto{\pgfqpoint{-0.000000in}{0.000000in}}%
\pgfpathlineto{\pgfqpoint{-0.048611in}{0.000000in}}%
\pgfusepath{stroke,fill}%
}%
\begin{pgfscope}%
\pgfsys@transformshift{0.000000in}{0.518824in}%
\pgfsys@useobject{currentmarker}{}%
\end{pgfscope}%
\end{pgfscope}%
\begin{pgfscope}%
\definecolor{textcolor}{rgb}{0.000000,0.000000,0.000000}%
\pgfsetstrokecolor{textcolor}%
\pgfsetfillcolor{textcolor}%
\pgftext[x=-0.148147in, y=0.489888in, left, base]{\color{textcolor}\rmfamily\fontsize{6.000000}{7.200000}\selectfont \(\displaystyle {3}\)}%
\end{pgfscope}%
\begin{pgfscope}%
\pgfsetbuttcap%
\pgfsetroundjoin%
\definecolor{currentfill}{rgb}{0.000000,0.000000,0.000000}%
\pgfsetfillcolor{currentfill}%
\pgfsetlinewidth{0.803000pt}%
\definecolor{currentstroke}{rgb}{0.000000,0.000000,0.000000}%
\pgfsetstrokecolor{currentstroke}%
\pgfsetdash{}{0pt}%
\pgfsys@defobject{currentmarker}{\pgfqpoint{-0.048611in}{0.000000in}}{\pgfqpoint{-0.000000in}{0.000000in}}{%
\pgfpathmoveto{\pgfqpoint{-0.000000in}{0.000000in}}%
\pgfpathlineto{\pgfqpoint{-0.048611in}{0.000000in}}%
\pgfusepath{stroke,fill}%
}%
\begin{pgfscope}%
\pgfsys@transformshift{0.000000in}{0.337647in}%
\pgfsys@useobject{currentmarker}{}%
\end{pgfscope}%
\end{pgfscope}%
\begin{pgfscope}%
\definecolor{textcolor}{rgb}{0.000000,0.000000,0.000000}%
\pgfsetstrokecolor{textcolor}%
\pgfsetfillcolor{textcolor}%
\pgftext[x=-0.148147in, y=0.308712in, left, base]{\color{textcolor}\rmfamily\fontsize{6.000000}{7.200000}\selectfont \(\displaystyle {4}\)}%
\end{pgfscope}%
\begin{pgfscope}%
\pgfsetrectcap%
\pgfsetmiterjoin%
\pgfsetlinewidth{0.803000pt}%
\definecolor{currentstroke}{rgb}{0.000000,0.000000,0.000000}%
\pgfsetstrokecolor{currentstroke}%
\pgfsetdash{}{0pt}%
\pgfpathmoveto{\pgfqpoint{0.000000in}{0.247059in}}%
\pgfpathlineto{\pgfqpoint{0.000000in}{1.152941in}}%
\pgfusepath{stroke}%
\end{pgfscope}%
\begin{pgfscope}%
\pgfsetrectcap%
\pgfsetmiterjoin%
\pgfsetlinewidth{0.803000pt}%
\definecolor{currentstroke}{rgb}{0.000000,0.000000,0.000000}%
\pgfsetstrokecolor{currentstroke}%
\pgfsetdash{}{0pt}%
\pgfpathmoveto{\pgfqpoint{0.905882in}{0.247059in}}%
\pgfpathlineto{\pgfqpoint{0.905882in}{1.152941in}}%
\pgfusepath{stroke}%
\end{pgfscope}%
\begin{pgfscope}%
\pgfsetrectcap%
\pgfsetmiterjoin%
\pgfsetlinewidth{0.803000pt}%
\definecolor{currentstroke}{rgb}{0.000000,0.000000,0.000000}%
\pgfsetstrokecolor{currentstroke}%
\pgfsetdash{}{0pt}%
\pgfpathmoveto{\pgfqpoint{0.000000in}{0.247059in}}%
\pgfpathlineto{\pgfqpoint{0.905882in}{0.247059in}}%
\pgfusepath{stroke}%
\end{pgfscope}%
\begin{pgfscope}%
\pgfsetrectcap%
\pgfsetmiterjoin%
\pgfsetlinewidth{0.803000pt}%
\definecolor{currentstroke}{rgb}{0.000000,0.000000,0.000000}%
\pgfsetstrokecolor{currentstroke}%
\pgfsetdash{}{0pt}%
\pgfpathmoveto{\pgfqpoint{0.000000in}{1.152941in}}%
\pgfpathlineto{\pgfqpoint{0.905882in}{1.152941in}}%
\pgfusepath{stroke}%
\end{pgfscope}%
\begin{pgfscope}%
\definecolor{textcolor}{rgb}{0.000000,0.000000,0.000000}%
\pgfsetstrokecolor{textcolor}%
\pgfsetfillcolor{textcolor}%
\pgftext[x=0.452941in,y=1.236275in,,base]{\color{textcolor}\rmfamily\fontsize{7.200000}{8.640000}\selectfont \(\displaystyle \cos \langle \mathbf p_i, \mathbf q_j \rangle\)}%
\end{pgfscope}%
\begin{pgfscope}%
\pgfsetbuttcap%
\pgfsetmiterjoin%
\definecolor{currentfill}{rgb}{1.000000,1.000000,1.000000}%
\pgfsetfillcolor{currentfill}%
\pgfsetlinewidth{0.000000pt}%
\definecolor{currentstroke}{rgb}{0.000000,0.000000,0.000000}%
\pgfsetstrokecolor{currentstroke}%
\pgfsetstrokeopacity{0.000000}%
\pgfsetdash{}{0pt}%
\pgfpathmoveto{\pgfqpoint{1.087059in}{0.247059in}}%
\pgfpathlineto{\pgfqpoint{1.992941in}{0.247059in}}%
\pgfpathlineto{\pgfqpoint{1.992941in}{1.152941in}}%
\pgfpathlineto{\pgfqpoint{1.087059in}{1.152941in}}%
\pgfpathlineto{\pgfqpoint{1.087059in}{0.247059in}}%
\pgfpathclose%
\pgfusepath{fill}%
\end{pgfscope}%
\begin{pgfscope}%
\pgfpathrectangle{\pgfqpoint{1.087059in}{0.247059in}}{\pgfqpoint{0.905882in}{0.905882in}}%
\pgfusepath{clip}%
\pgfsys@transformshift{1.087059in}{0.247059in}%
\pgftext[left,bottom]{\includegraphics[interpolate=true,width=0.910000in,height=0.910000in]{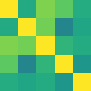}}%
\end{pgfscope}%
\begin{pgfscope}%
\pgfsetbuttcap%
\pgfsetroundjoin%
\definecolor{currentfill}{rgb}{0.000000,0.000000,0.000000}%
\pgfsetfillcolor{currentfill}%
\pgfsetlinewidth{0.803000pt}%
\definecolor{currentstroke}{rgb}{0.000000,0.000000,0.000000}%
\pgfsetstrokecolor{currentstroke}%
\pgfsetdash{}{0pt}%
\pgfsys@defobject{currentmarker}{\pgfqpoint{0.000000in}{-0.048611in}}{\pgfqpoint{0.000000in}{0.000000in}}{%
\pgfpathmoveto{\pgfqpoint{0.000000in}{0.000000in}}%
\pgfpathlineto{\pgfqpoint{0.000000in}{-0.048611in}}%
\pgfusepath{stroke,fill}%
}%
\begin{pgfscope}%
\pgfsys@transformshift{1.177647in}{0.247059in}%
\pgfsys@useobject{currentmarker}{}%
\end{pgfscope}%
\end{pgfscope}%
\begin{pgfscope}%
\definecolor{textcolor}{rgb}{0.000000,0.000000,0.000000}%
\pgfsetstrokecolor{textcolor}%
\pgfsetfillcolor{textcolor}%
\pgftext[x=1.177647in,y=0.149837in,,top]{\color{textcolor}\rmfamily\fontsize{6.000000}{7.200000}\selectfont \(\displaystyle {0}\)}%
\end{pgfscope}%
\begin{pgfscope}%
\pgfsetbuttcap%
\pgfsetroundjoin%
\definecolor{currentfill}{rgb}{0.000000,0.000000,0.000000}%
\pgfsetfillcolor{currentfill}%
\pgfsetlinewidth{0.803000pt}%
\definecolor{currentstroke}{rgb}{0.000000,0.000000,0.000000}%
\pgfsetstrokecolor{currentstroke}%
\pgfsetdash{}{0pt}%
\pgfsys@defobject{currentmarker}{\pgfqpoint{0.000000in}{-0.048611in}}{\pgfqpoint{0.000000in}{0.000000in}}{%
\pgfpathmoveto{\pgfqpoint{0.000000in}{0.000000in}}%
\pgfpathlineto{\pgfqpoint{0.000000in}{-0.048611in}}%
\pgfusepath{stroke,fill}%
}%
\begin{pgfscope}%
\pgfsys@transformshift{1.540000in}{0.247059in}%
\pgfsys@useobject{currentmarker}{}%
\end{pgfscope}%
\end{pgfscope}%
\begin{pgfscope}%
\definecolor{textcolor}{rgb}{0.000000,0.000000,0.000000}%
\pgfsetstrokecolor{textcolor}%
\pgfsetfillcolor{textcolor}%
\pgftext[x=1.540000in,y=0.149837in,,top]{\color{textcolor}\rmfamily\fontsize{6.000000}{7.200000}\selectfont \(\displaystyle {2}\)}%
\end{pgfscope}%
\begin{pgfscope}%
\pgfsetbuttcap%
\pgfsetroundjoin%
\definecolor{currentfill}{rgb}{0.000000,0.000000,0.000000}%
\pgfsetfillcolor{currentfill}%
\pgfsetlinewidth{0.803000pt}%
\definecolor{currentstroke}{rgb}{0.000000,0.000000,0.000000}%
\pgfsetstrokecolor{currentstroke}%
\pgfsetdash{}{0pt}%
\pgfsys@defobject{currentmarker}{\pgfqpoint{0.000000in}{-0.048611in}}{\pgfqpoint{0.000000in}{0.000000in}}{%
\pgfpathmoveto{\pgfqpoint{0.000000in}{0.000000in}}%
\pgfpathlineto{\pgfqpoint{0.000000in}{-0.048611in}}%
\pgfusepath{stroke,fill}%
}%
\begin{pgfscope}%
\pgfsys@transformshift{1.902353in}{0.247059in}%
\pgfsys@useobject{currentmarker}{}%
\end{pgfscope}%
\end{pgfscope}%
\begin{pgfscope}%
\definecolor{textcolor}{rgb}{0.000000,0.000000,0.000000}%
\pgfsetstrokecolor{textcolor}%
\pgfsetfillcolor{textcolor}%
\pgftext[x=1.902353in,y=0.149837in,,top]{\color{textcolor}\rmfamily\fontsize{6.000000}{7.200000}\selectfont \(\displaystyle {4}\)}%
\end{pgfscope}%
\begin{pgfscope}%
\pgfsetbuttcap%
\pgfsetroundjoin%
\definecolor{currentfill}{rgb}{0.000000,0.000000,0.000000}%
\pgfsetfillcolor{currentfill}%
\pgfsetlinewidth{0.803000pt}%
\definecolor{currentstroke}{rgb}{0.000000,0.000000,0.000000}%
\pgfsetstrokecolor{currentstroke}%
\pgfsetdash{}{0pt}%
\pgfsys@defobject{currentmarker}{\pgfqpoint{-0.048611in}{0.000000in}}{\pgfqpoint{-0.000000in}{0.000000in}}{%
\pgfpathmoveto{\pgfqpoint{-0.000000in}{0.000000in}}%
\pgfpathlineto{\pgfqpoint{-0.048611in}{0.000000in}}%
\pgfusepath{stroke,fill}%
}%
\begin{pgfscope}%
\pgfsys@transformshift{1.087059in}{1.062353in}%
\pgfsys@useobject{currentmarker}{}%
\end{pgfscope}%
\end{pgfscope}%
\begin{pgfscope}%
\definecolor{textcolor}{rgb}{0.000000,0.000000,0.000000}%
\pgfsetstrokecolor{textcolor}%
\pgfsetfillcolor{textcolor}%
\pgftext[x=0.938911in, y=1.033418in, left, base]{\color{textcolor}\rmfamily\fontsize{6.000000}{7.200000}\selectfont \(\displaystyle {0}\)}%
\end{pgfscope}%
\begin{pgfscope}%
\pgfsetbuttcap%
\pgfsetroundjoin%
\definecolor{currentfill}{rgb}{0.000000,0.000000,0.000000}%
\pgfsetfillcolor{currentfill}%
\pgfsetlinewidth{0.803000pt}%
\definecolor{currentstroke}{rgb}{0.000000,0.000000,0.000000}%
\pgfsetstrokecolor{currentstroke}%
\pgfsetdash{}{0pt}%
\pgfsys@defobject{currentmarker}{\pgfqpoint{-0.048611in}{0.000000in}}{\pgfqpoint{-0.000000in}{0.000000in}}{%
\pgfpathmoveto{\pgfqpoint{-0.000000in}{0.000000in}}%
\pgfpathlineto{\pgfqpoint{-0.048611in}{0.000000in}}%
\pgfusepath{stroke,fill}%
}%
\begin{pgfscope}%
\pgfsys@transformshift{1.087059in}{0.881176in}%
\pgfsys@useobject{currentmarker}{}%
\end{pgfscope}%
\end{pgfscope}%
\begin{pgfscope}%
\definecolor{textcolor}{rgb}{0.000000,0.000000,0.000000}%
\pgfsetstrokecolor{textcolor}%
\pgfsetfillcolor{textcolor}%
\pgftext[x=0.938911in, y=0.852241in, left, base]{\color{textcolor}\rmfamily\fontsize{6.000000}{7.200000}\selectfont \(\displaystyle {1}\)}%
\end{pgfscope}%
\begin{pgfscope}%
\pgfsetbuttcap%
\pgfsetroundjoin%
\definecolor{currentfill}{rgb}{0.000000,0.000000,0.000000}%
\pgfsetfillcolor{currentfill}%
\pgfsetlinewidth{0.803000pt}%
\definecolor{currentstroke}{rgb}{0.000000,0.000000,0.000000}%
\pgfsetstrokecolor{currentstroke}%
\pgfsetdash{}{0pt}%
\pgfsys@defobject{currentmarker}{\pgfqpoint{-0.048611in}{0.000000in}}{\pgfqpoint{-0.000000in}{0.000000in}}{%
\pgfpathmoveto{\pgfqpoint{-0.000000in}{0.000000in}}%
\pgfpathlineto{\pgfqpoint{-0.048611in}{0.000000in}}%
\pgfusepath{stroke,fill}%
}%
\begin{pgfscope}%
\pgfsys@transformshift{1.087059in}{0.700000in}%
\pgfsys@useobject{currentmarker}{}%
\end{pgfscope}%
\end{pgfscope}%
\begin{pgfscope}%
\definecolor{textcolor}{rgb}{0.000000,0.000000,0.000000}%
\pgfsetstrokecolor{textcolor}%
\pgfsetfillcolor{textcolor}%
\pgftext[x=0.938911in, y=0.671065in, left, base]{\color{textcolor}\rmfamily\fontsize{6.000000}{7.200000}\selectfont \(\displaystyle {2}\)}%
\end{pgfscope}%
\begin{pgfscope}%
\pgfsetbuttcap%
\pgfsetroundjoin%
\definecolor{currentfill}{rgb}{0.000000,0.000000,0.000000}%
\pgfsetfillcolor{currentfill}%
\pgfsetlinewidth{0.803000pt}%
\definecolor{currentstroke}{rgb}{0.000000,0.000000,0.000000}%
\pgfsetstrokecolor{currentstroke}%
\pgfsetdash{}{0pt}%
\pgfsys@defobject{currentmarker}{\pgfqpoint{-0.048611in}{0.000000in}}{\pgfqpoint{-0.000000in}{0.000000in}}{%
\pgfpathmoveto{\pgfqpoint{-0.000000in}{0.000000in}}%
\pgfpathlineto{\pgfqpoint{-0.048611in}{0.000000in}}%
\pgfusepath{stroke,fill}%
}%
\begin{pgfscope}%
\pgfsys@transformshift{1.087059in}{0.518824in}%
\pgfsys@useobject{currentmarker}{}%
\end{pgfscope}%
\end{pgfscope}%
\begin{pgfscope}%
\definecolor{textcolor}{rgb}{0.000000,0.000000,0.000000}%
\pgfsetstrokecolor{textcolor}%
\pgfsetfillcolor{textcolor}%
\pgftext[x=0.938911in, y=0.489888in, left, base]{\color{textcolor}\rmfamily\fontsize{6.000000}{7.200000}\selectfont \(\displaystyle {3}\)}%
\end{pgfscope}%
\begin{pgfscope}%
\pgfsetbuttcap%
\pgfsetroundjoin%
\definecolor{currentfill}{rgb}{0.000000,0.000000,0.000000}%
\pgfsetfillcolor{currentfill}%
\pgfsetlinewidth{0.803000pt}%
\definecolor{currentstroke}{rgb}{0.000000,0.000000,0.000000}%
\pgfsetstrokecolor{currentstroke}%
\pgfsetdash{}{0pt}%
\pgfsys@defobject{currentmarker}{\pgfqpoint{-0.048611in}{0.000000in}}{\pgfqpoint{-0.000000in}{0.000000in}}{%
\pgfpathmoveto{\pgfqpoint{-0.000000in}{0.000000in}}%
\pgfpathlineto{\pgfqpoint{-0.048611in}{0.000000in}}%
\pgfusepath{stroke,fill}%
}%
\begin{pgfscope}%
\pgfsys@transformshift{1.087059in}{0.337647in}%
\pgfsys@useobject{currentmarker}{}%
\end{pgfscope}%
\end{pgfscope}%
\begin{pgfscope}%
\definecolor{textcolor}{rgb}{0.000000,0.000000,0.000000}%
\pgfsetstrokecolor{textcolor}%
\pgfsetfillcolor{textcolor}%
\pgftext[x=0.938911in, y=0.308712in, left, base]{\color{textcolor}\rmfamily\fontsize{6.000000}{7.200000}\selectfont \(\displaystyle {4}\)}%
\end{pgfscope}%
\begin{pgfscope}%
\pgfsetrectcap%
\pgfsetmiterjoin%
\pgfsetlinewidth{0.803000pt}%
\definecolor{currentstroke}{rgb}{0.000000,0.000000,0.000000}%
\pgfsetstrokecolor{currentstroke}%
\pgfsetdash{}{0pt}%
\pgfpathmoveto{\pgfqpoint{1.087059in}{0.247059in}}%
\pgfpathlineto{\pgfqpoint{1.087059in}{1.152941in}}%
\pgfusepath{stroke}%
\end{pgfscope}%
\begin{pgfscope}%
\pgfsetrectcap%
\pgfsetmiterjoin%
\pgfsetlinewidth{0.803000pt}%
\definecolor{currentstroke}{rgb}{0.000000,0.000000,0.000000}%
\pgfsetstrokecolor{currentstroke}%
\pgfsetdash{}{0pt}%
\pgfpathmoveto{\pgfqpoint{1.992941in}{0.247059in}}%
\pgfpathlineto{\pgfqpoint{1.992941in}{1.152941in}}%
\pgfusepath{stroke}%
\end{pgfscope}%
\begin{pgfscope}%
\pgfsetrectcap%
\pgfsetmiterjoin%
\pgfsetlinewidth{0.803000pt}%
\definecolor{currentstroke}{rgb}{0.000000,0.000000,0.000000}%
\pgfsetstrokecolor{currentstroke}%
\pgfsetdash{}{0pt}%
\pgfpathmoveto{\pgfqpoint{1.087059in}{0.247059in}}%
\pgfpathlineto{\pgfqpoint{1.992941in}{0.247059in}}%
\pgfusepath{stroke}%
\end{pgfscope}%
\begin{pgfscope}%
\pgfsetrectcap%
\pgfsetmiterjoin%
\pgfsetlinewidth{0.803000pt}%
\definecolor{currentstroke}{rgb}{0.000000,0.000000,0.000000}%
\pgfsetstrokecolor{currentstroke}%
\pgfsetdash{}{0pt}%
\pgfpathmoveto{\pgfqpoint{1.087059in}{1.152941in}}%
\pgfpathlineto{\pgfqpoint{1.992941in}{1.152941in}}%
\pgfusepath{stroke}%
\end{pgfscope}%
\begin{pgfscope}%
\definecolor{textcolor}{rgb}{0.000000,0.000000,0.000000}%
\pgfsetstrokecolor{textcolor}%
\pgfsetfillcolor{textcolor}%
\pgftext[x=1.540000in,y=1.236275in,,base]{\color{textcolor}\rmfamily\fontsize{7.200000}{8.640000}\selectfont \(\displaystyle \cos \langle \mathbf p_i, \mathbf p_j \rangle\)}%
\end{pgfscope}%
\begin{pgfscope}%
\pgfsetbuttcap%
\pgfsetmiterjoin%
\definecolor{currentfill}{rgb}{1.000000,1.000000,1.000000}%
\pgfsetfillcolor{currentfill}%
\pgfsetlinewidth{0.000000pt}%
\definecolor{currentstroke}{rgb}{0.000000,0.000000,0.000000}%
\pgfsetstrokecolor{currentstroke}%
\pgfsetstrokeopacity{0.000000}%
\pgfsetdash{}{0pt}%
\pgfpathmoveto{\pgfqpoint{2.174118in}{0.247059in}}%
\pgfpathlineto{\pgfqpoint{3.080000in}{0.247059in}}%
\pgfpathlineto{\pgfqpoint{3.080000in}{1.152941in}}%
\pgfpathlineto{\pgfqpoint{2.174118in}{1.152941in}}%
\pgfpathlineto{\pgfqpoint{2.174118in}{0.247059in}}%
\pgfpathclose%
\pgfusepath{fill}%
\end{pgfscope}%
\begin{pgfscope}%
\pgfpathrectangle{\pgfqpoint{2.174118in}{0.247059in}}{\pgfqpoint{0.905882in}{0.905882in}}%
\pgfusepath{clip}%
\pgfsys@transformshift{2.170000in}{0.250000in}%
\pgftext[left,bottom]{\includegraphics[interpolate=true,width=0.900000in,height=0.910000in]{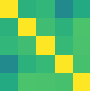}}%
\end{pgfscope}%
\begin{pgfscope}%
\pgfsetbuttcap%
\pgfsetroundjoin%
\definecolor{currentfill}{rgb}{0.000000,0.000000,0.000000}%
\pgfsetfillcolor{currentfill}%
\pgfsetlinewidth{0.803000pt}%
\definecolor{currentstroke}{rgb}{0.000000,0.000000,0.000000}%
\pgfsetstrokecolor{currentstroke}%
\pgfsetdash{}{0pt}%
\pgfsys@defobject{currentmarker}{\pgfqpoint{0.000000in}{-0.048611in}}{\pgfqpoint{0.000000in}{0.000000in}}{%
\pgfpathmoveto{\pgfqpoint{0.000000in}{0.000000in}}%
\pgfpathlineto{\pgfqpoint{0.000000in}{-0.048611in}}%
\pgfusepath{stroke,fill}%
}%
\begin{pgfscope}%
\pgfsys@transformshift{2.264706in}{0.247059in}%
\pgfsys@useobject{currentmarker}{}%
\end{pgfscope}%
\end{pgfscope}%
\begin{pgfscope}%
\definecolor{textcolor}{rgb}{0.000000,0.000000,0.000000}%
\pgfsetstrokecolor{textcolor}%
\pgfsetfillcolor{textcolor}%
\pgftext[x=2.264706in,y=0.149837in,,top]{\color{textcolor}\rmfamily\fontsize{6.000000}{7.200000}\selectfont \(\displaystyle {0}\)}%
\end{pgfscope}%
\begin{pgfscope}%
\pgfsetbuttcap%
\pgfsetroundjoin%
\definecolor{currentfill}{rgb}{0.000000,0.000000,0.000000}%
\pgfsetfillcolor{currentfill}%
\pgfsetlinewidth{0.803000pt}%
\definecolor{currentstroke}{rgb}{0.000000,0.000000,0.000000}%
\pgfsetstrokecolor{currentstroke}%
\pgfsetdash{}{0pt}%
\pgfsys@defobject{currentmarker}{\pgfqpoint{0.000000in}{-0.048611in}}{\pgfqpoint{0.000000in}{0.000000in}}{%
\pgfpathmoveto{\pgfqpoint{0.000000in}{0.000000in}}%
\pgfpathlineto{\pgfqpoint{0.000000in}{-0.048611in}}%
\pgfusepath{stroke,fill}%
}%
\begin{pgfscope}%
\pgfsys@transformshift{2.627059in}{0.247059in}%
\pgfsys@useobject{currentmarker}{}%
\end{pgfscope}%
\end{pgfscope}%
\begin{pgfscope}%
\definecolor{textcolor}{rgb}{0.000000,0.000000,0.000000}%
\pgfsetstrokecolor{textcolor}%
\pgfsetfillcolor{textcolor}%
\pgftext[x=2.627059in,y=0.149837in,,top]{\color{textcolor}\rmfamily\fontsize{6.000000}{7.200000}\selectfont \(\displaystyle {2}\)}%
\end{pgfscope}%
\begin{pgfscope}%
\pgfsetbuttcap%
\pgfsetroundjoin%
\definecolor{currentfill}{rgb}{0.000000,0.000000,0.000000}%
\pgfsetfillcolor{currentfill}%
\pgfsetlinewidth{0.803000pt}%
\definecolor{currentstroke}{rgb}{0.000000,0.000000,0.000000}%
\pgfsetstrokecolor{currentstroke}%
\pgfsetdash{}{0pt}%
\pgfsys@defobject{currentmarker}{\pgfqpoint{0.000000in}{-0.048611in}}{\pgfqpoint{0.000000in}{0.000000in}}{%
\pgfpathmoveto{\pgfqpoint{0.000000in}{0.000000in}}%
\pgfpathlineto{\pgfqpoint{0.000000in}{-0.048611in}}%
\pgfusepath{stroke,fill}%
}%
\begin{pgfscope}%
\pgfsys@transformshift{2.989412in}{0.247059in}%
\pgfsys@useobject{currentmarker}{}%
\end{pgfscope}%
\end{pgfscope}%
\begin{pgfscope}%
\definecolor{textcolor}{rgb}{0.000000,0.000000,0.000000}%
\pgfsetstrokecolor{textcolor}%
\pgfsetfillcolor{textcolor}%
\pgftext[x=2.989412in,y=0.149837in,,top]{\color{textcolor}\rmfamily\fontsize{6.000000}{7.200000}\selectfont \(\displaystyle {4}\)}%
\end{pgfscope}%
\begin{pgfscope}%
\pgfsetbuttcap%
\pgfsetroundjoin%
\definecolor{currentfill}{rgb}{0.000000,0.000000,0.000000}%
\pgfsetfillcolor{currentfill}%
\pgfsetlinewidth{0.803000pt}%
\definecolor{currentstroke}{rgb}{0.000000,0.000000,0.000000}%
\pgfsetstrokecolor{currentstroke}%
\pgfsetdash{}{0pt}%
\pgfsys@defobject{currentmarker}{\pgfqpoint{-0.048611in}{0.000000in}}{\pgfqpoint{-0.000000in}{0.000000in}}{%
\pgfpathmoveto{\pgfqpoint{-0.000000in}{0.000000in}}%
\pgfpathlineto{\pgfqpoint{-0.048611in}{0.000000in}}%
\pgfusepath{stroke,fill}%
}%
\begin{pgfscope}%
\pgfsys@transformshift{2.174118in}{1.062353in}%
\pgfsys@useobject{currentmarker}{}%
\end{pgfscope}%
\end{pgfscope}%
\begin{pgfscope}%
\definecolor{textcolor}{rgb}{0.000000,0.000000,0.000000}%
\pgfsetstrokecolor{textcolor}%
\pgfsetfillcolor{textcolor}%
\pgftext[x=2.025970in, y=1.033418in, left, base]{\color{textcolor}\rmfamily\fontsize{6.000000}{7.200000}\selectfont \(\displaystyle {0}\)}%
\end{pgfscope}%
\begin{pgfscope}%
\pgfsetbuttcap%
\pgfsetroundjoin%
\definecolor{currentfill}{rgb}{0.000000,0.000000,0.000000}%
\pgfsetfillcolor{currentfill}%
\pgfsetlinewidth{0.803000pt}%
\definecolor{currentstroke}{rgb}{0.000000,0.000000,0.000000}%
\pgfsetstrokecolor{currentstroke}%
\pgfsetdash{}{0pt}%
\pgfsys@defobject{currentmarker}{\pgfqpoint{-0.048611in}{0.000000in}}{\pgfqpoint{-0.000000in}{0.000000in}}{%
\pgfpathmoveto{\pgfqpoint{-0.000000in}{0.000000in}}%
\pgfpathlineto{\pgfqpoint{-0.048611in}{0.000000in}}%
\pgfusepath{stroke,fill}%
}%
\begin{pgfscope}%
\pgfsys@transformshift{2.174118in}{0.881176in}%
\pgfsys@useobject{currentmarker}{}%
\end{pgfscope}%
\end{pgfscope}%
\begin{pgfscope}%
\definecolor{textcolor}{rgb}{0.000000,0.000000,0.000000}%
\pgfsetstrokecolor{textcolor}%
\pgfsetfillcolor{textcolor}%
\pgftext[x=2.025970in, y=0.852241in, left, base]{\color{textcolor}\rmfamily\fontsize{6.000000}{7.200000}\selectfont \(\displaystyle {1}\)}%
\end{pgfscope}%
\begin{pgfscope}%
\pgfsetbuttcap%
\pgfsetroundjoin%
\definecolor{currentfill}{rgb}{0.000000,0.000000,0.000000}%
\pgfsetfillcolor{currentfill}%
\pgfsetlinewidth{0.803000pt}%
\definecolor{currentstroke}{rgb}{0.000000,0.000000,0.000000}%
\pgfsetstrokecolor{currentstroke}%
\pgfsetdash{}{0pt}%
\pgfsys@defobject{currentmarker}{\pgfqpoint{-0.048611in}{0.000000in}}{\pgfqpoint{-0.000000in}{0.000000in}}{%
\pgfpathmoveto{\pgfqpoint{-0.000000in}{0.000000in}}%
\pgfpathlineto{\pgfqpoint{-0.048611in}{0.000000in}}%
\pgfusepath{stroke,fill}%
}%
\begin{pgfscope}%
\pgfsys@transformshift{2.174118in}{0.700000in}%
\pgfsys@useobject{currentmarker}{}%
\end{pgfscope}%
\end{pgfscope}%
\begin{pgfscope}%
\definecolor{textcolor}{rgb}{0.000000,0.000000,0.000000}%
\pgfsetstrokecolor{textcolor}%
\pgfsetfillcolor{textcolor}%
\pgftext[x=2.025970in, y=0.671065in, left, base]{\color{textcolor}\rmfamily\fontsize{6.000000}{7.200000}\selectfont \(\displaystyle {2}\)}%
\end{pgfscope}%
\begin{pgfscope}%
\pgfsetbuttcap%
\pgfsetroundjoin%
\definecolor{currentfill}{rgb}{0.000000,0.000000,0.000000}%
\pgfsetfillcolor{currentfill}%
\pgfsetlinewidth{0.803000pt}%
\definecolor{currentstroke}{rgb}{0.000000,0.000000,0.000000}%
\pgfsetstrokecolor{currentstroke}%
\pgfsetdash{}{0pt}%
\pgfsys@defobject{currentmarker}{\pgfqpoint{-0.048611in}{0.000000in}}{\pgfqpoint{-0.000000in}{0.000000in}}{%
\pgfpathmoveto{\pgfqpoint{-0.000000in}{0.000000in}}%
\pgfpathlineto{\pgfqpoint{-0.048611in}{0.000000in}}%
\pgfusepath{stroke,fill}%
}%
\begin{pgfscope}%
\pgfsys@transformshift{2.174118in}{0.518824in}%
\pgfsys@useobject{currentmarker}{}%
\end{pgfscope}%
\end{pgfscope}%
\begin{pgfscope}%
\definecolor{textcolor}{rgb}{0.000000,0.000000,0.000000}%
\pgfsetstrokecolor{textcolor}%
\pgfsetfillcolor{textcolor}%
\pgftext[x=2.025970in, y=0.489888in, left, base]{\color{textcolor}\rmfamily\fontsize{6.000000}{7.200000}\selectfont \(\displaystyle {3}\)}%
\end{pgfscope}%
\begin{pgfscope}%
\pgfsetbuttcap%
\pgfsetroundjoin%
\definecolor{currentfill}{rgb}{0.000000,0.000000,0.000000}%
\pgfsetfillcolor{currentfill}%
\pgfsetlinewidth{0.803000pt}%
\definecolor{currentstroke}{rgb}{0.000000,0.000000,0.000000}%
\pgfsetstrokecolor{currentstroke}%
\pgfsetdash{}{0pt}%
\pgfsys@defobject{currentmarker}{\pgfqpoint{-0.048611in}{0.000000in}}{\pgfqpoint{-0.000000in}{0.000000in}}{%
\pgfpathmoveto{\pgfqpoint{-0.000000in}{0.000000in}}%
\pgfpathlineto{\pgfqpoint{-0.048611in}{0.000000in}}%
\pgfusepath{stroke,fill}%
}%
\begin{pgfscope}%
\pgfsys@transformshift{2.174118in}{0.337647in}%
\pgfsys@useobject{currentmarker}{}%
\end{pgfscope}%
\end{pgfscope}%
\begin{pgfscope}%
\definecolor{textcolor}{rgb}{0.000000,0.000000,0.000000}%
\pgfsetstrokecolor{textcolor}%
\pgfsetfillcolor{textcolor}%
\pgftext[x=2.025970in, y=0.308712in, left, base]{\color{textcolor}\rmfamily\fontsize{6.000000}{7.200000}\selectfont \(\displaystyle {4}\)}%
\end{pgfscope}%
\begin{pgfscope}%
\pgfsetrectcap%
\pgfsetmiterjoin%
\pgfsetlinewidth{0.803000pt}%
\definecolor{currentstroke}{rgb}{0.000000,0.000000,0.000000}%
\pgfsetstrokecolor{currentstroke}%
\pgfsetdash{}{0pt}%
\pgfpathmoveto{\pgfqpoint{2.174118in}{0.247059in}}%
\pgfpathlineto{\pgfqpoint{2.174118in}{1.152941in}}%
\pgfusepath{stroke}%
\end{pgfscope}%
\begin{pgfscope}%
\pgfsetrectcap%
\pgfsetmiterjoin%
\pgfsetlinewidth{0.803000pt}%
\definecolor{currentstroke}{rgb}{0.000000,0.000000,0.000000}%
\pgfsetstrokecolor{currentstroke}%
\pgfsetdash{}{0pt}%
\pgfpathmoveto{\pgfqpoint{3.080000in}{0.247059in}}%
\pgfpathlineto{\pgfqpoint{3.080000in}{1.152941in}}%
\pgfusepath{stroke}%
\end{pgfscope}%
\begin{pgfscope}%
\pgfsetrectcap%
\pgfsetmiterjoin%
\pgfsetlinewidth{0.803000pt}%
\definecolor{currentstroke}{rgb}{0.000000,0.000000,0.000000}%
\pgfsetstrokecolor{currentstroke}%
\pgfsetdash{}{0pt}%
\pgfpathmoveto{\pgfqpoint{2.174118in}{0.247059in}}%
\pgfpathlineto{\pgfqpoint{3.080000in}{0.247059in}}%
\pgfusepath{stroke}%
\end{pgfscope}%
\begin{pgfscope}%
\pgfsetrectcap%
\pgfsetmiterjoin%
\pgfsetlinewidth{0.803000pt}%
\definecolor{currentstroke}{rgb}{0.000000,0.000000,0.000000}%
\pgfsetstrokecolor{currentstroke}%
\pgfsetdash{}{0pt}%
\pgfpathmoveto{\pgfqpoint{2.174118in}{1.152941in}}%
\pgfpathlineto{\pgfqpoint{3.080000in}{1.152941in}}%
\pgfusepath{stroke}%
\end{pgfscope}%
\begin{pgfscope}%
\definecolor{textcolor}{rgb}{0.000000,0.000000,0.000000}%
\pgfsetstrokecolor{textcolor}%
\pgfsetfillcolor{textcolor}%
\pgftext[x=2.627059in,y=1.236275in,,base]{\color{textcolor}\rmfamily\fontsize{7.200000}{8.640000}\selectfont \(\displaystyle \cos \langle \mathbf q_i, \mathbf q_j \rangle\)}%
\end{pgfscope}%
\begin{pgfscope}%
\pgfsetbuttcap%
\pgfsetmiterjoin%
\definecolor{currentfill}{rgb}{1.000000,1.000000,1.000000}%
\pgfsetfillcolor{currentfill}%
\pgfsetlinewidth{0.000000pt}%
\definecolor{currentstroke}{rgb}{0.000000,0.000000,0.000000}%
\pgfsetstrokecolor{currentstroke}%
\pgfsetstrokeopacity{0.000000}%
\pgfsetdash{}{0pt}%
\pgfpathmoveto{\pgfqpoint{3.272500in}{0.350000in}}%
\pgfpathlineto{\pgfqpoint{3.307500in}{0.350000in}}%
\pgfpathlineto{\pgfqpoint{3.307500in}{1.050000in}}%
\pgfpathlineto{\pgfqpoint{3.272500in}{1.050000in}}%
\pgfpathlineto{\pgfqpoint{3.272500in}{0.350000in}}%
\pgfpathclose%
\pgfusepath{fill}%
\end{pgfscope}%
\begin{pgfscope}%
\pgfpathrectangle{\pgfqpoint{3.272500in}{0.350000in}}{\pgfqpoint{0.035000in}{0.700000in}}%
\pgfusepath{clip}%
\pgfsetbuttcap%
\pgfsetmiterjoin%
\definecolor{currentfill}{rgb}{1.000000,1.000000,1.000000}%
\pgfsetfillcolor{currentfill}%
\pgfsetlinewidth{0.010037pt}%
\definecolor{currentstroke}{rgb}{1.000000,1.000000,1.000000}%
\pgfsetstrokecolor{currentstroke}%
\pgfsetdash{}{0pt}%
\pgfusepath{stroke,fill}%
\end{pgfscope}%
\begin{pgfscope}%
\pgfsys@transformshift{3.270000in}{0.350000in}%
\pgftext[left,bottom]{\includegraphics[interpolate=true,width=0.040000in,height=0.700000in]{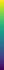}}%
\end{pgfscope}%
\begin{pgfscope}%
\pgfsetbuttcap%
\pgfsetroundjoin%
\definecolor{currentfill}{rgb}{0.000000,0.000000,0.000000}%
\pgfsetfillcolor{currentfill}%
\pgfsetlinewidth{0.803000pt}%
\definecolor{currentstroke}{rgb}{0.000000,0.000000,0.000000}%
\pgfsetstrokecolor{currentstroke}%
\pgfsetdash{}{0pt}%
\pgfsys@defobject{currentmarker}{\pgfqpoint{0.000000in}{0.000000in}}{\pgfqpoint{0.048611in}{0.000000in}}{%
\pgfpathmoveto{\pgfqpoint{0.000000in}{0.000000in}}%
\pgfpathlineto{\pgfqpoint{0.048611in}{0.000000in}}%
\pgfusepath{stroke,fill}%
}%
\begin{pgfscope}%
\pgfsys@transformshift{3.307500in}{0.350000in}%
\pgfsys@useobject{currentmarker}{}%
\end{pgfscope}%
\end{pgfscope}%
\begin{pgfscope}%
\definecolor{textcolor}{rgb}{0.000000,0.000000,0.000000}%
\pgfsetstrokecolor{textcolor}%
\pgfsetfillcolor{textcolor}%
\pgftext[x=3.404722in, y=0.321065in, left, base]{\color{textcolor}\rmfamily\fontsize{6.000000}{7.200000}\selectfont \(\displaystyle {0.00}\)}%
\end{pgfscope}%
\begin{pgfscope}%
\pgfsetbuttcap%
\pgfsetroundjoin%
\definecolor{currentfill}{rgb}{0.000000,0.000000,0.000000}%
\pgfsetfillcolor{currentfill}%
\pgfsetlinewidth{0.803000pt}%
\definecolor{currentstroke}{rgb}{0.000000,0.000000,0.000000}%
\pgfsetstrokecolor{currentstroke}%
\pgfsetdash{}{0pt}%
\pgfsys@defobject{currentmarker}{\pgfqpoint{0.000000in}{0.000000in}}{\pgfqpoint{0.048611in}{0.000000in}}{%
\pgfpathmoveto{\pgfqpoint{0.000000in}{0.000000in}}%
\pgfpathlineto{\pgfqpoint{0.048611in}{0.000000in}}%
\pgfusepath{stroke,fill}%
}%
\begin{pgfscope}%
\pgfsys@transformshift{3.307500in}{0.525000in}%
\pgfsys@useobject{currentmarker}{}%
\end{pgfscope}%
\end{pgfscope}%
\begin{pgfscope}%
\definecolor{textcolor}{rgb}{0.000000,0.000000,0.000000}%
\pgfsetstrokecolor{textcolor}%
\pgfsetfillcolor{textcolor}%
\pgftext[x=3.404722in, y=0.496065in, left, base]{\color{textcolor}\rmfamily\fontsize{6.000000}{7.200000}\selectfont \(\displaystyle {0.25}\)}%
\end{pgfscope}%
\begin{pgfscope}%
\pgfsetbuttcap%
\pgfsetroundjoin%
\definecolor{currentfill}{rgb}{0.000000,0.000000,0.000000}%
\pgfsetfillcolor{currentfill}%
\pgfsetlinewidth{0.803000pt}%
\definecolor{currentstroke}{rgb}{0.000000,0.000000,0.000000}%
\pgfsetstrokecolor{currentstroke}%
\pgfsetdash{}{0pt}%
\pgfsys@defobject{currentmarker}{\pgfqpoint{0.000000in}{0.000000in}}{\pgfqpoint{0.048611in}{0.000000in}}{%
\pgfpathmoveto{\pgfqpoint{0.000000in}{0.000000in}}%
\pgfpathlineto{\pgfqpoint{0.048611in}{0.000000in}}%
\pgfusepath{stroke,fill}%
}%
\begin{pgfscope}%
\pgfsys@transformshift{3.307500in}{0.700000in}%
\pgfsys@useobject{currentmarker}{}%
\end{pgfscope}%
\end{pgfscope}%
\begin{pgfscope}%
\definecolor{textcolor}{rgb}{0.000000,0.000000,0.000000}%
\pgfsetstrokecolor{textcolor}%
\pgfsetfillcolor{textcolor}%
\pgftext[x=3.404722in, y=0.671065in, left, base]{\color{textcolor}\rmfamily\fontsize{6.000000}{7.200000}\selectfont \(\displaystyle {0.50}\)}%
\end{pgfscope}%
\begin{pgfscope}%
\pgfsetbuttcap%
\pgfsetroundjoin%
\definecolor{currentfill}{rgb}{0.000000,0.000000,0.000000}%
\pgfsetfillcolor{currentfill}%
\pgfsetlinewidth{0.803000pt}%
\definecolor{currentstroke}{rgb}{0.000000,0.000000,0.000000}%
\pgfsetstrokecolor{currentstroke}%
\pgfsetdash{}{0pt}%
\pgfsys@defobject{currentmarker}{\pgfqpoint{0.000000in}{0.000000in}}{\pgfqpoint{0.048611in}{0.000000in}}{%
\pgfpathmoveto{\pgfqpoint{0.000000in}{0.000000in}}%
\pgfpathlineto{\pgfqpoint{0.048611in}{0.000000in}}%
\pgfusepath{stroke,fill}%
}%
\begin{pgfscope}%
\pgfsys@transformshift{3.307500in}{0.875000in}%
\pgfsys@useobject{currentmarker}{}%
\end{pgfscope}%
\end{pgfscope}%
\begin{pgfscope}%
\definecolor{textcolor}{rgb}{0.000000,0.000000,0.000000}%
\pgfsetstrokecolor{textcolor}%
\pgfsetfillcolor{textcolor}%
\pgftext[x=3.404722in, y=0.846065in, left, base]{\color{textcolor}\rmfamily\fontsize{6.000000}{7.200000}\selectfont \(\displaystyle {0.75}\)}%
\end{pgfscope}%
\begin{pgfscope}%
\pgfsetbuttcap%
\pgfsetroundjoin%
\definecolor{currentfill}{rgb}{0.000000,0.000000,0.000000}%
\pgfsetfillcolor{currentfill}%
\pgfsetlinewidth{0.803000pt}%
\definecolor{currentstroke}{rgb}{0.000000,0.000000,0.000000}%
\pgfsetstrokecolor{currentstroke}%
\pgfsetdash{}{0pt}%
\pgfsys@defobject{currentmarker}{\pgfqpoint{0.000000in}{0.000000in}}{\pgfqpoint{0.048611in}{0.000000in}}{%
\pgfpathmoveto{\pgfqpoint{0.000000in}{0.000000in}}%
\pgfpathlineto{\pgfqpoint{0.048611in}{0.000000in}}%
\pgfusepath{stroke,fill}%
}%
\begin{pgfscope}%
\pgfsys@transformshift{3.307500in}{1.050000in}%
\pgfsys@useobject{currentmarker}{}%
\end{pgfscope}%
\end{pgfscope}%
\begin{pgfscope}%
\definecolor{textcolor}{rgb}{0.000000,0.000000,0.000000}%
\pgfsetstrokecolor{textcolor}%
\pgfsetfillcolor{textcolor}%
\pgftext[x=3.404722in, y=1.021065in, left, base]{\color{textcolor}\rmfamily\fontsize{6.000000}{7.200000}\selectfont \(\displaystyle {1.00}\)}%
\end{pgfscope}%
\begin{pgfscope}%
\pgfsetrectcap%
\pgfsetmiterjoin%
\pgfsetlinewidth{0.803000pt}%
\definecolor{currentstroke}{rgb}{0.000000,0.000000,0.000000}%
\pgfsetstrokecolor{currentstroke}%
\pgfsetdash{}{0pt}%
\pgfpathmoveto{\pgfqpoint{3.272500in}{0.350000in}}%
\pgfpathlineto{\pgfqpoint{3.290000in}{0.350000in}}%
\pgfpathlineto{\pgfqpoint{3.307500in}{0.350000in}}%
\pgfpathlineto{\pgfqpoint{3.307500in}{1.050000in}}%
\pgfpathlineto{\pgfqpoint{3.290000in}{1.050000in}}%
\pgfpathlineto{\pgfqpoint{3.272500in}{1.050000in}}%
\pgfpathlineto{\pgfqpoint{3.272500in}{0.350000in}}%
\pgfpathclose%
\pgfusepath{stroke}%
\end{pgfscope}%
\end{pgfpicture}%
\makeatother%
\endgroup%